\documentclass[letterpaper, 10 pt, conference]{ieeeconf}  
\let\labelindent\relax
\IEEEoverridecommandlockouts 
\usepackage{algorithm}
\usepackage[utf8]{inputenc}
\usepackage{algpseudocode}
\usepackage{amsmath}

\usepackage[T1]{fontenc}
\usepackage[inline]{enumitem}
\usepackage[table,xcdraw]{xcolor}
\usepackage{multirow}
\usepackage{color}
\usepackage{tabularray}
\usepackage{pgfplots}
\usepackage{tikz}
\usepackage{subcaption}
\pgfplotsset{compat=newest}
\usepackage{xcolor}
\usepackage{setspace}
\usepackage{amssymb}
\usepackage{mathtools}
\usepackage[hidelinks]{hyperref}

\title{\LARGE \bf Accurate Pose Estimation Using Contact Manifold Sampling for Safe Peg-in-Hole Insertion of Complex Geometries} 

\author{Abhay Negi$^{1}$, Omey M. Manyar$^{1}$, Dhanush K. Penmetsa$^{1}$, and Satyandra K. Gupta$^{1}$
\thanks{This work was supported by NASA Space Technology Graduate Research Opportunities Award (80NSSC24K1380).}%
\thanks{$^{1}$Realization of Robotic Systems Lab, University of Southern California, Los Angeles, CA, USA. Address all correspondence to \href{mailto:guptask@usc.edu}{guptask@usc.edu}}}

\begin{document}

\maketitle
\thispagestyle{empty}
\pagestyle{empty}

\begin{abstract}
Robotic assembly of complex, non-convex geometries with tight clearances remains a challenging problem, demanding precise state estimation for successful insertion. In this work, we propose a novel framework that relies solely on contact states to estimate the full $SE(3)$ pose of a peg relative to a hole. Our method constructs an online submanifold of contact states through primitive motions with just 6 seconds of online execution, subsequently mapping it to an offline contact manifold for precise pose estimation. We demonstrate that without such state estimation, robots risk jamming and excessive force application, potentially causing damage. We evaluate our approach on five industrially relevant, complex geometries with 0.1 to 1.0 mm clearances, achieving a 96.7\% success rate-a 6× improvement over primitive-based insertion without state estimation. Additionally, we analyze insertion forces, and overall insertion times, showing our method significantly reduces the average wrench, enabling safer and more efficient assembly.

\end{abstract}


\section{INTRODUCTION}

\noindent{The peg-in-hole insertion task is one of the most fundamental problems in robotics. In the realm of contact-rich manipulation and assembly, insertion-based tasks are often framed as peg-in-hole problems. Over the decades, research in this area has achieved remarkable progress, with many studies successfully performing insertion tasks under high uncertainty, due to advancements in state estimation and deep learning-based methods \cite{jiang_review_2022, shen_review_2023}.}

Most prior work has focused on simple, convex geometries such as square or circular pegs; however, complex, non-convex geometries - such as gear assemblies, automobile chassis, and turbine blades - remain largely understudied and present unique challenges for successful task execution. To quantify geometrical complexity, metrics such as the deviation from the convex hull of the cross-sectional profile may be used \cite{brinkhoff_measuring_1995}. 
Assembly of complex geometries is difficult as small pose misalignment often causes obstructive contacts, often leading to insertion failure. Even for simple geometries, misalignment issues are exacerbated with low clearance-to-diameter/length ratios and high-friction surfaces \cite{whitney_quasi-static_1982, jiang_review_2022}. 

\begin{figure}[ht]
    \centering
    \includegraphics[width=1.0\linewidth]{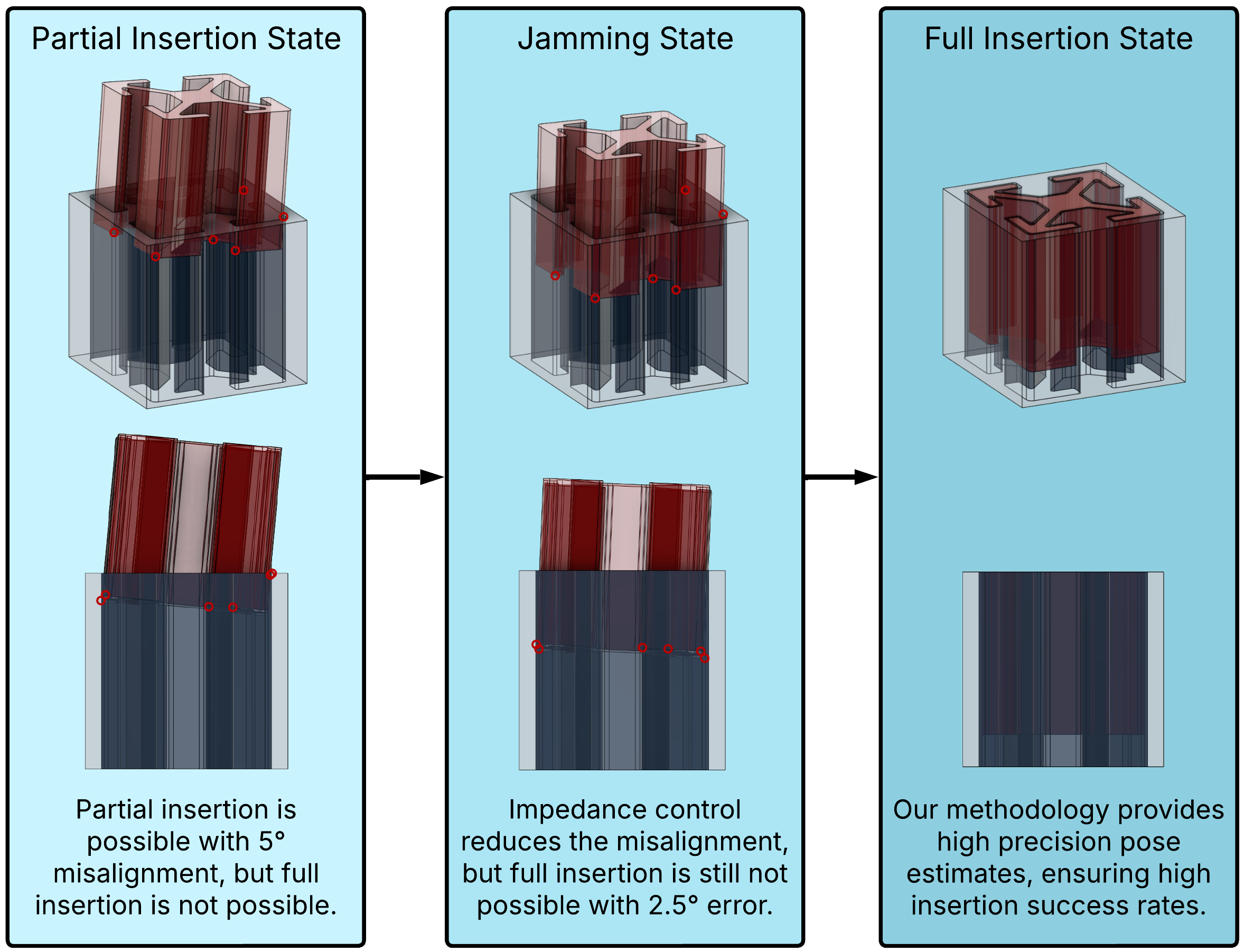}
    \caption{The primary challenge of assembling complex geometries with tight clearances arises from misalignments that result in contacts (red circles), obstructing further insertion.} 
    \label{fig:jamming}
\end{figure}

State estimation is crucial for successful insertion, as failures often stem from misalignment (Refer Fig. \ref{fig:jamming}). While impedance control can compensate for pose errors, the compensation due to its active compliance is inadequate for complex, tight-clearance geometries. As shown in Section \ref{sec:results}, inaccurate pose estimates lead to jamming and low success rates. Even when successful, impedance control alone exerts higher forces, risking part damage.

Alternatively, computer vision has been widely used for pose estimation; however, for high-precision assembly of complex geometries, such methods struggle due to a lack of visual features in partially inserted states. Feature/key point matching methods, although promising, are particularly sensitive to noise, especially in low-textured geometries. Similarly, learning-based approaches require large real-world datasets or techniques such as domain randomization to remain robust to variations in lighting and backgrounds.

Wrench-based techniques are widely used for contact state and pose estimation but typically lack full observability of the $SE(3)$ pose and are highly sensitive to measurement errors. Additionally, distinct poses can produce similar wrench values, making pose inference an ill-posed inverse problem due to non-uniqueness. While multiple measurements can help resolve this, instability remains a challenge. Wrench sensors are prone to noise and calibration drift, often requiring substantial applied force to achieve a sufficient signal-to-noise ratio.

In contrast, a robotic arm’s proprioceptive sensors provide reliable and precise joint position measurements, enabling end-effector pose estimation with 0.01 to 0.1 mm and deg scale precision. Additionally, force-based contact detection remains highly reliable despite noise and calibration errors. Our approach leverages these measurements for precise pose estimation, ensuring robust and reliable assembly. While vision and wrench data can be integrated, as discussed in Section \ref{sec:conclusions}, we focus on demonstrating the effectiveness of pose measurements and contact detection alone.

Our methodology employs a contact manifold that represents all feasible poses where contact occurs between the peg and hole. This manifold is unique to each geometry and possesses distinct features that allow effective sampling and alignment with a limited set of observations. Due to the advantages of joint position measurements and contact detection, we can construct an accurate and precise contact manifold. Additionally, the manifold is robust to errors, with measurement inaccuracies leading to localized, non-propagating errors. Finally, the contact manifold can be safely sampled without risking part damage, as contact detection only requires a low amount of force.

In this work, we present a method that safely samples the contact manifold during execution, enabling precise pose estimation for reliable and robust insertion of complex geometries with tight clearances. Our approach relies solely on contact detection, making it particularly effective in scenarios with occlusions or noisy measurements. This contact-based method ensures robust geometric information regardless of environmental conditions. We evaluate it on five industrially relevant, complex geometries with maximum clearances between 0.1 mm and 1.0 mm, achieving a 96.7\% success rate. The main contributions of this work are:

\begin{itemize}
    \item A novel paradigm and methodology for performing peg-in-hole insertion of complex geometry using only kinematic measurements and force-based contact detection   
    
    \item A general-purpose representation of contact between two bodies - the contact manifold, and a sampling-based method to feasibly construct a representation of the manifold   
    
    \item An algorithm to compute the $SE(3)$ relative transform between a reference contact manifold and an observed contact submanifold
\end{itemize}

\section{RELATED WORKS} 
\label{sec:related-works}

\noindent\textbf{State Estimation-Based Insertion:} Peg-in-hole insertion has been a longstanding challenge in robotics \cite{bruyninckx_peg--hole_1995, dutre1996contact, tsaprounis1998contact}, with prior work primarily taking one of two approaches: (1) crafting a direct insertion policy \cite{zhang_icra_2022} or (2) performing explicit state estimation \cite{pankert_learning_2023}. In both cases, state estimation plays a role, albeit implicitly in the first and explicitly in the second. Our work closely aligns with the latter. A common approach in state estimation is to explicitly compute the pose of the hole relative to the robot, leveraging advanced sensing modalities such as force \cite{lee_polyfit_2023}, vision \cite{haugaard_case_2022, liang_icra_2022, negi_autonomous_nodate}, and tactile feedback \cite{wu20241}. Recent advances in learning-based frameworks have enabled multimodal fusion for improved estimation \cite{lee2020making}; however, these methods often demand extensive real-world data or sophisticated sim-to-real transfer techniques \cite{lee_polyfit_2023, lammle_iros_2022, pmlr-v229-zhang23e}. Moreover, they remain susceptible to environmental variations and have been largely limited to simple geometries, with limited exploration of non-convex and complex shapes. An alternative approach focuses on contact state estimation \cite{jin2021contact, lee_ral_2022, pankert_learning_2023, whitney_quasi-static_1982}, which models the sequence of interactions between the peg and the hole. This line of research leverages the making and breaking of contacts to infer state transitions, aiding in robust insertion. While both end-to-end learning-based and model-driven methods have been explored, few have effectively tackled complex geometries \cite{song_automated_2014} and their impact on contact states. Our work builds upon this direction, aiming to extend contact state estimation to more intricate insertion scenarios.

\noindent \textbf{Reinforcement Learning: } Reinforcement learning (RL) has emerged as a powerful approach for robotic assembly due to its ability to learn complex dynamics that traditional controllers struggle to model. Early applications of RL to industrial insertion tasks by \cite{schoettler_deep_2020} demonstrated success with visual inputs and natural rewards by combining RL with prior information from classical controllers or demonstrations. However, these approaches faced limitations in handling multi-stage tasks, complex geometries, and adapting to previous inaccuracies. To address efficiency and safety concerns, \cite{schoettler_meta-reinforcement_2020, ankile2025imitationrefinementresidual} introduced meta-learning to solve simulated insertion tasks and then rapidly adapt to real-world conditions with minimal trials, while \cite{zhao_offline_2022} further reduced data requirements through offline meta-RL using demonstrations and prior task data. Recent advances have focused on bridging the sim-to-real gap for contact-rich manipulation. \cite{tang_industreal_2023} developed sim-aware policy updates and signed-distance-field rewards to enable successful transfer to real-world assembly tasks. However, minimal focus has been on complex geometries and tight clearance insertion. 

\noindent\textbf{Complex Geometry \& Jamming: }Geometries with significant complexity which have been studied in robotic assembly largely fall within two categories - multi-pegs and complex planar parts. Controller \cite{zhang_force_2017} and RL-based \cite{hou_fuzzy_2022} methods have been designed for insertion of 2-pin and 3-pin geometries \cite{zhang_force_2017, zhang_jamming_2019, fei_assembly_nodate}. Methods for automatically generating robust assembly sequences \cite{stemmer_robust_2006, stemmer_analytical_2007, giordano_robotic_2008} as well as force-controllers for insertion \cite{song_automated_2014, song_guidance_2016} have been developed for complex planar geometries. The majority of prior work which addresses jamming takes the analytical approach of enumerating all jamming cases and deriving force and moment relations with pose. Jamming has been analyzed for cylindrical geometry \cite{whitney_quasi-static_1982}, rectangular geometry \cite{caine_assembly_1989}, and 2-pin and 3-pin geometries \cite{sathirakul_jamming_1998, fei_jamming_2005, zhang_jamming_2019, fei_assembly_nodate}. To date, the most complex geometry for which jamming analysis has been performed is the circular-rectangular geometry \cite{wu_robot_2022}, which has 44 unique contact states. Jamming analysis of geometries with higher complexity quickly becomes intractable due to the combinatorial growth in number of unique contact states. Apart from jamming analysis, jamming has also been addressed on cylindrical geometry with wrench-based pose estimation \cite{son_neuralfuzzy_2001} and axial vibration \cite{kilikevicius_dynamic_2011}. 

\section{PROBLEM FORMULATION}

\subsection{Preliminaries and Assumptions}
\noindent We consider an agent, $\mathcal{A}$, tasked with manipulating an object or peg, denoted as $p \in \mathcal{P}$, where $\mathcal{P}$ represents a set of objects with complex, non-convex geometries. Each peg $p$ has a corresponding hole geometry $h$ such that the clearance between them is defined by $C(p, h) < \epsilon$, where $C(\cdot, \cdot)$ is a function that quantifies the clearance between $p$ and $h$.

The agent $\mathcal{A}$ has access to a set of predefined controllers, $\Omega$, and can observe the in-hand pose of the peg $p$, represented by a rigid transformation $^{\mathcal{A}}T_p \in SE(3)$. The objective of $\mathcal{A}$ is to perform an insertion task, $\mathcal{T}$, by estimating the relative pose of the hole with respect to the peg, denoted as $^pT_h$. At each time step $t$, the agent receives a noisy observation $\mathcal{O}_t$ (Refer Section. \ref{ssec:math-formulation}). Based on this observation, the agent computes an estimate of $^\mathcal{A}T_h$ and executes an action to align $p$ with $h$ by invoking appropriate controllers in $\Omega$, ultimately achieving a successful insertion.

\noindent Assumptions: We define the key assumptions for the task $\mathcal{T}$ as follows:
\begin{itemize}[leftmargin=10pt]
\item In-hand pose of the peg, $^\mathcal{A}T_p$, is known  without any uncertainty
\item Clearance threshold $\epsilon$ is between 0.1 and 1.0 mm
\item $\mathcal{A}$ starts the task $\mathcal{T}$ from a partially inserted configuration, meaning the peg $p$ is always interacting with the hole $h$
\end{itemize}

\subsection{Mathematical Formulation}
\label{ssec:math-formulation}
\noindent To estimate the relative pose $^hT_p$, we define a function $\Psi$ that maps the set of observations $\mathcal{O}$ to an estimated state of $h$ with respect to $p$. Prior work (see Section \ref{sec:related-works}) and our own experiments suggest that contact states of $p$, characterized by a subset of its configuration space $\mathcal{C}_p \in SE(3)$ where $p$ and $h$ are in contact, provide valuable information for modeling $\Psi$. We hypothesize the existence of a reference contact manifold $\mathcal{M} \subset \mathcal{C}_p$ in $p$'s configuration space, where $p$ and $h$ remain in contact. Furthermore, we define $\mathcal{M}$ in the frame of reference of the hole, $h$.

Given that $\mathcal{A}$ has access to this reference contact manifold $\mathcal{M}$ as prior knowledge, it can collect structured observations during the execution of task $\mathcal{T}$. Specifically, $\mathcal{A}$ performs deterministic motions to record states where contact occurs, which can be identified using an external force-torque sensor. Contact is detected when the measured external force magnitude satisfies $|F_{ext}| > \epsilon_f$, where $\epsilon_f$ is the experimentally determined minimum force threshold for contact. These contact observations, denoted as $\mathcal{O}_{contact}$, are initially represented in $\mathcal{A}$'s frame of reference. Since $\mathcal{O}_{contact}$ belongs to the contact manifold $\mathcal{M}$, estimating $^hT_p$ reduces to determining the mapping between the submanifold represented by $\mathcal{O}_{contact}$ in $\mathcal{A}$'s frame and the contact manifold $\mathcal{M}$ in $h$'s frame. Therefore, we define the function $\Psi$ as: $\Psi(\mathcal{M}, \mathcal{O}_{contact}) \mapsto \ ^\mathcal{A}T_h$. Our objective is to solve for the rigid transformation between the contact submanifold observed by $\mathcal{A}$ and the contact manifold $\mathcal{M}$ defined in the hole’s frame of reference.

\section{METHODOLOGY}

\begin{figure*}[t!]
    \centering
    \makebox[\textwidth][c]{%
        \includegraphics[width=0.88\textwidth]{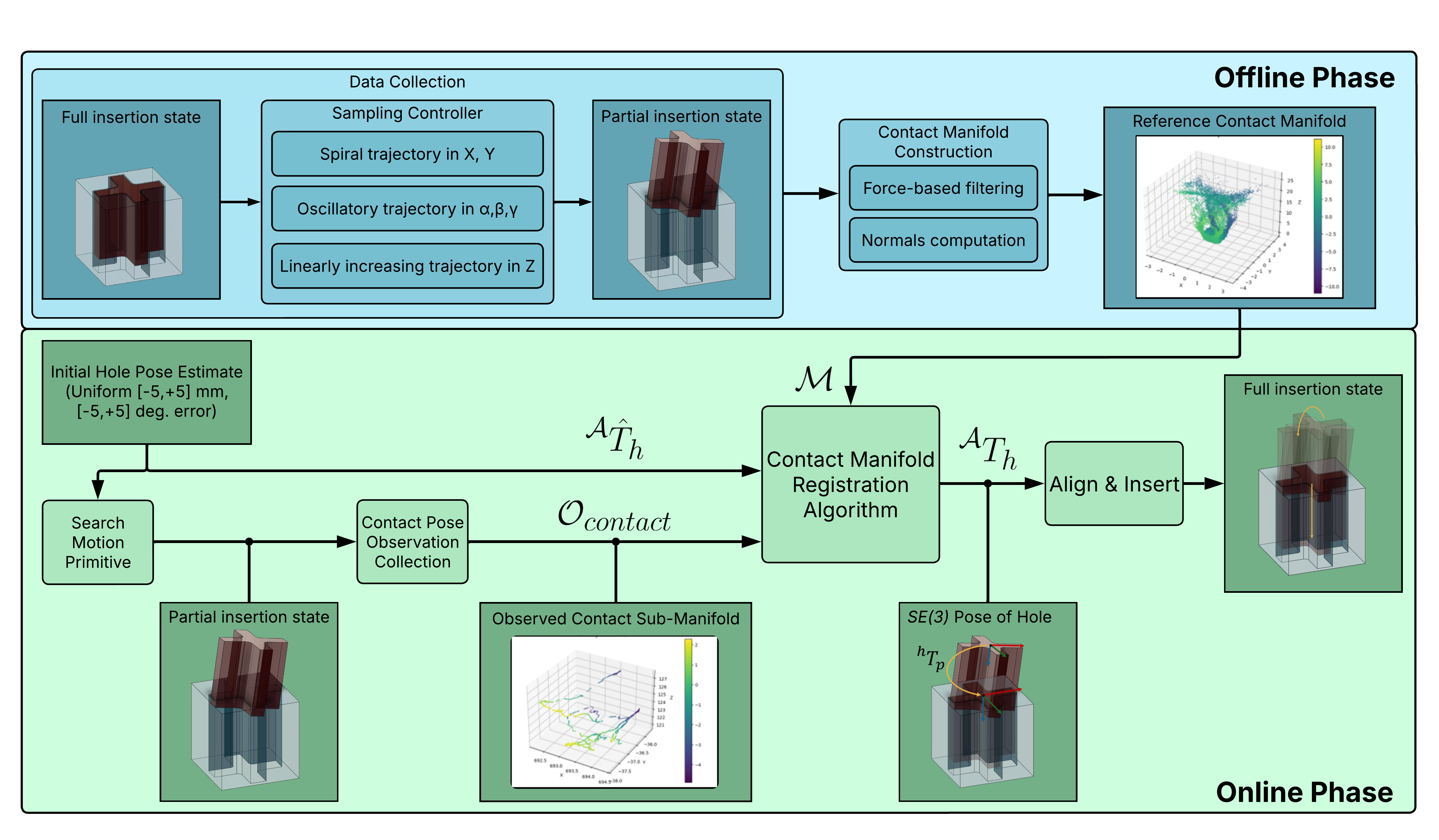}
    }
    \caption{The Overview of our methodology. During the offline phase, a sampling controller is used to construct the reference contact manifold, $\mathcal{M}$. During the online phase, the system is provided an initial hole pose estimate with significant error. The search motion primitive uses this estimate to achieve a partial insertion state. Next, contact pose observation collection is performed by perturbing the peg pose, resulting in the observed contact submanifold, $\mathcal{O}_{contact}$. The contact manifold registration algorithm uses $\mathcal{O}_{contact}$, $\mathcal{M}$, and the initial hole pose estimate to compute a precise pose of the hole with respect to the robot. The system then uses this pose to align the peg and finally insert to achieve the full insertion state.}
    \label{fig:system-overview}   
\end{figure*}

\bigskip

\begin{algorithm*}[t]
\footnotesize
\setstretch{1.15}
\caption{Contact Manifold Registration Algorithm} 
\label{alg:pose_icp}
\begin{algorithmic}[1]
    \State \textbf{Data:} $\{(^HT_p, \hat{n}_q)\}$ \algorithmiccomment{Reference contact manifold point-normal pairs}

    \State \textbf{Input:} $\{^\mathcal{A}T_p\}$, $^\mathcal{A}T_{h,0}$ \algorithmiccomment{Observed contact submanifold \& initial estimate} 
    
    \State $\{^{h,0}T_{p}\} \gets \left( ^{\mathcal{A}}T_{h,0} \right)^{-1} \otimes \{ ^{\mathcal{A}}T_{p} \}$ \algorithmiccomment{Initial alignment} \label{line:initial alignment}
    
    \For {$t = 0$ \textbf{to} $N-1$} 

        \For {$^{h,t}T_{p,i} \in \{^{h,t}T_p\}$} 

            \State $(^{H}T_{p,i}, \hat{n}_{q,i}) \gets \text{nearest neighbor of } ^{h,t}T_{p,i}$ \text{from} $\{^{H}T_{p}\}$  \algorithmiccomment{Find nearest neighbor from reference manifold} 

            \State $\delta_i \gets \left( ^{h,t}T_{p,i}\right)^{-1} \otimes \left( ^HT_{p,i} \right) $ \algorithmiccomment{Compute point-to-point misalignment} 

            \State $\mathcal{B} \gets \text{matrix of k nearest neighbors of } ^{h,t}T_{p,i} \text{ from } \{^{h,t}T_{p}\} $ \algorithmiccomment{Find k nearest neighbors from observations} 

            \State $\hat{n}_{p,i} \gets \text{singular vector of min. singular value of } SVD \left( COV\left(\mathcal{B} - \left(^{h,t}T_{p,i}\right)\right) \right) $ \algorithmiccomment{Estimate observation normal} \label{line:observation normal}

            \State $^t\Delta_i \gets \frac{1}{2} \left[ \left( \delta_i \cdot \hat{n}_{q,i} \right)\hat{n}_{q,i} + \left( \delta_i \cdot \hat{n}_{p,i} \right) \hat{n}_{p,i} \right] $ \algorithmiccomment{Compute symmetric point-to-plane misalignment} \label{line:symmetric point-to-plane}     
            
        \EndFor

        \State $ ^{h,t}T_{h,t+1} \gets SE(3) \text{ mean pose of } \{ ^t\Delta \} $ \algorithmiccomment{Compute aggregate pose update from misalignments} 

        \State $  ^\mathcal{A}T_{h,t+1}  \gets \left( ^\mathcal{A}T_{h,t} \right) \otimes \left( ^{h,t}T_{h,t+1} \right)$   \algorithmiccomment{Update hole pose estimate} 
        
        \State $\{ ^{h,t+1}T_{p} \} \gets \left( ^{h,t}T_{h,t+1} \right)^{-1} \otimes \{ ^{h,t}T_{p} \} $ \algorithmiccomment{Update observations}
                    
    \EndFor 

    \State \textbf{return} $^{\mathcal{A}}T_{h,N}$
    
    \end{algorithmic}
\end{algorithm*}
\setstretch{1}

\subsection{Overview}

\noindent To solve for the function $\Psi$, we propose an approach illustrated in Fig. \ref{fig:system-overview}. Our methodology consists of two stages: a reference offline contact manifold generation phase and an online state estimation phase. In the offline phase, our goal is to construct the reference contact manifold $\mathcal{M}$. Specifically, $\mathcal{M}$ is a 6-D manifold comprising of the poses represented by $x,y,z$ position and Euler angles $\alpha,\beta,\gamma$, where the peg and hole are in contact. To construct $\mathcal{M}$, we fix the hole geometry at a known pose relative to the robot ($\mathcal{A}$) and command the robot to explore poses where contact occurs between the peg and hole. This process is detailed in Section \ref{ssec:reference-manifold-generation}.

The online phase focuses on estimating the relative transformation $^\mathcal{A}T_h$ by collecting contact observations $\mathcal{O}_{contact}$ and computing the corresponding contact submanifold. Our method builds on the iterative closest point (ICP) algorithm to iteratively estimate $^\mathcal{A}T_h$, aligning the observed contact submanifold with  the reference contact manifold $\mathcal{M}$ (see Section \ref{ssec:submanifold-generation}). Our experimental setup consists of a KUKA LBR iiwa 14 R820 robot, which operates in two control modes ($\Omega$): (1) Joint Position Control and (2) Cartesian Impedance Control. The peg geometries are rigidly attached to the robot's end-effector (see Section \ref{ssec:cell-setup}). Additionally, KUKA LBR iiwa includes joint torque sensing capability, which provides us with end-effector wrench information. This information is used to identify the contact poses (Refer Section. \ref{ssec:submanifold-generation}). We assume an initial hole pose uncertainty of up to 5 mm in position and up to $5^\circ$ in orientation. The online phase process flow is depicted in Fig. \ref{fig:system-overview}. The process concludes upon successful insertion.

\subsection{Reference Contact Manifold Generation}
\label{ssec:reference-manifold-generation}
\noindent Contact-based insertion strategies rely on a reference model of possible contact configurations. We define this as the reference contact manifold, $\mathcal{M}$, capturing the geometric relationship between the peg $p$ and hole $h$ during contact. This manifold helps estimate the relative pose during assembly. 

\noindent\textbf{Geometry-Induced Manifold Structure: }
The structure of $\mathcal{M}$ is determined by the geometries of the interacting parts. Each distinct geometry creates a unique "fingerprint" in the 6D pose space that encodes its contact characteristics. For instance, square pegs generate manifolds with sharp transitions at corners and linear segments along edges, while cylindrical pegs produce radially symmetric, continuous manifolds. This correspondence between physical geometry and manifold structure is fundamental to our approach, enabling pose estimation through manifold registration.

\noindent\textbf{Sampling Methodology}
We represent the contact manifold as a collection of discrete poses $^{h}T_{p}$ where contact occurs. Our sampling utilizes a primitive combining spiral trajectories in translation with oscillatory motions in angular dimensions. These motions are executed with varying amplitude parameters depending on clearances and insertion depth to prevent part damage. Low-stiffness impedance control enables natural compliance during contact, and we only record poses where contact force exceeds the threshold ($F_{ext} > \epsilon_f$).

We incorporate random perturbations in the exploration trajectories to mitigate sampling bias. Since any method that would use the contact manifold for state estimation would depend on geometric alignment, the distribution of samples significantly impacts performance. Registration algorithms operate by iteratively minimizing the distance between corresponding observation and reference points. Non-uniform spatial coverage introduces systematic biases because the optimization process disproportionately weighs regions with higher point density. In such cases, densely sampled areas dominate the error metric, even if their geometric features are less discriminative for alignment. For example, oversampling non-unique features (like planar surfaces) while undersampling geometrically distinct features (like corners or edges) biases registration toward overfitting the less informative regions, potentially misaligning the critical structural elements that would better constrain the transformation.

The number of samples represents a critical trade-off between estimation accuracy and computational efficiency. Any method using the reference manifold $\mathcal{M}$ and observation submanifold $\mathcal{O}_{contact}$ for state estimation would scale quadratically with sample count, primarily due to correspondence matching requirements. Most estimation techniques rely on k-nearest neighbor search algorithms, which we further discuss in Section. \ref{ssec:pose-estimation}. We found that for contact manifolds exceeding $10^6$ points, computational costs become prohibitive, necessitating downsampling. However, excessive downsampling risks losing critical features, leading to overly sparse manifolds ($<10^3$ points) lacking sufficient constraints for reliable transformation estimation, particularly in feature-poor regions of the manifold. Conversely, insufficient downsampling in the presence of biased sampling may paradoxically lead to local minima trapping by providing numerous false correspondences \cite{Holz}. Our experiments indicate optimal performance in the $10^4$-$10^5$ range. 

\subsection{Contact Observation Collection}
\label{ssec:submanifold-generation}
\noindent After constructing $\mathcal{M}$ during the offline phase, we collect contact observations $\mathcal{O}_{contact} = \{^{\mathcal{A}}T_p\}$ at test time to estimate the unknown hole pose. Our collection strategy employs spiral trajectories in X and Y dimensions combined with oscillatory movements for angular orientations $\alpha$, $\beta$, and $\gamma$, constrained by a fixed time budget $t_{obs}$. In the first phase, we execute wider amplitude motions until a target insertion depth is reached. Once at this depth, we transition to smaller amplitude motions, taking advantage of the naturally constrained configuration space at deeper insertion levels. This focused sampling at greater insertion depths provides more discriminative geometric features for accurate registration as contact points become more tightly clustered around the true alignment axis.

Throughout both phases, the peg's pose $^{\mathcal{A}}T_{p}$ and wrench measurements are continuously recorded, with contact observations filtered using a force threshold: $\mathcal{O}_{contact} = \{ ^{\mathcal{A}}T_{p} \mid F_{ext} > F_{threshold}\}$. These filtered observations form a contact submanifold in $\mathcal{A}$'s frame, which is subsequently registered against the reference manifold $\mathcal{M}$ to estimate $^{\mathcal{A}}T_h$. Given the temporal continuity of these observations, we adopt a temporal downsampling procedure to a fixed number of points, which simultaneously increases variance in the sample set and reduces computational cost for registration. This structured observation approach provides us with informative observations, a contact submanifold with distinct features that can lead to better estimates in pose estimation while avoiding the jamming that often occurs during blind search strategies. With these contact observations collected, our next step is to determine the transformation between this observation submanifold and the reference manifold.

\subsection{Pose Estimation} 
\label{ssec:pose-estimation}
\noindent{The $SE(3)$ pose of hole $h$ with respect to the agent $\mathcal{A}$ is estimated by utilizing the contact manifold registration algorithm outlined in Algorithm \ref{alg:pose_icp}. The algorithm iteratively determines the $SE(3)$ pose which aligns the observed contact submanifold, $\mathcal{O}_{contact}$ with the reference contact manifold, $\mathcal{M}$. The contact observations and reference contact manifold are each a point cloud, where each point is a contact pose represented using a position vector and Euler angles, i.e. $T = \left[ x,y,z,\alpha,\beta,\gamma\right]^T \in \mathbb{R}^6$. Similar to ICP in $\mathbb{R}^3$, the observations are iteratively aligned with the manifold by determining correspondences between the two sets, computing misalignments, aligning the observations to the manifold, and updating the pose estimate. }

For each point of the reference contact manifold, the normal vector is computed - this defines a set of tuples, $\{\left( ^HT_p, \hat{n}_q \right)\}$, where $^HT_p$ is a feasible contact pose between the peg and hole and $\hat{n}_q$ is the local normal vector of $\{ ^HT_p \}$. The local normal vector is computed using singular value decomposition of a small neighborhood of each point. The observed contact submanifold is a set of poses of the peg with respect to the agent, $\{ ^\mathcal{A}T_p \}$. Normal vectors are also computed for the observed submanifold, but since they are not $SE(3)$ invariant, they are computed each loop iteration, as shown in Line \ref{line:observation normal} of Algorithm \ref{alg:pose_icp}. The initial estimate, $^\mathcal{A}T_{h,0}$, is used to perform a coarse initial alignment of the set of observations to the hole frame, as shown in Line \ref{line:initial alignment}. 

Within each iteration of the loop, the single nearest neighbor from the reference contact manifold is found for each observation. This is done by querying a k-d tree defined using the Euclidean distance metric. The index of the corresponding nearest neighbor is also used to identify the corresponding normal vector $\hat{n}_{q,i}$. The reference-to-observation, point-to-point misalignment, $\delta_i$, is computed by computing the $SE(3)$ relative pose between the corresponding observation and reference point. Next, the k nearest neighbors from the observations is found for each observation, which are used to compute the local normal vector of the observation. This is done by centering the neighbors about the observation point, computing the covariance matrix, and performing singular value decomposition. The singular vector corresponding to the minimum singular value is an estimate of the local normal vector of the observations at the specified point. 

The point-to-point misalignments, $\delta_i$, and local normal vectors from the reference manifold and observation submanifold, $\hat{n}_{q,i}$ and $\hat{n}_{p,i}$, are then used to compute the symmetric point-to-plane misalignment, as shown in Line \ref{line:symmetric point-to-plane}. The symmetric point-to-plane misalignment represents the average of the misalignment between the point of the observation to the plane local to the corresponding point of the reference contact manifold, and the misalignment between the point of the reference to the plane local to the corresponding point of the observation. 

After computing misalignments at each observation point, the aggregate misalignment is obtained as the $SE(3)$ mean pose, which is used to update the pose estimate and observations. This loop continues iteratively until the maximum iterations are reached. To ensure robustness to poor initializations and local minima, multiple optimization loops are run in parallel with initial poses sampled from a Gaussian distribution. The final estimate is obtained by averaging the resulting poses using the $SE(3)$ mean.

\subsection{Insertion}
\label{ssec:insertion}
\noindent After estimating $^{\mathcal{A}}T_h$ using Algorithm \ref{alg:pose_icp}, we execute a controlled insertion motion that accounts for residual alignment errors. Our strategy consists of a minor retract to prevent jamming, followed by an in-plane alignment step at constant insertion depth under high impedance control, and finally the insertion motion with optimized compliance parameters. The insertion is considered successful when $|z_{final} - z_{success}| \leq \epsilon_{success}$, where $z_{final}$ and $z_{success}$ are the z-values of the peg with respect to the hole at the final and success configurations, respectively, and $\epsilon_{success}$ is the success threshold.

\section{EXPERIMENTS \& RESULTS} 

\subsection{Cell Setup and Experiments} 
\label{ssec:cell-setup}
\begin{figure}
    \centering    
    \makebox[0.42\textwidth][c]{%
        \includegraphics[width=1.0\linewidth]{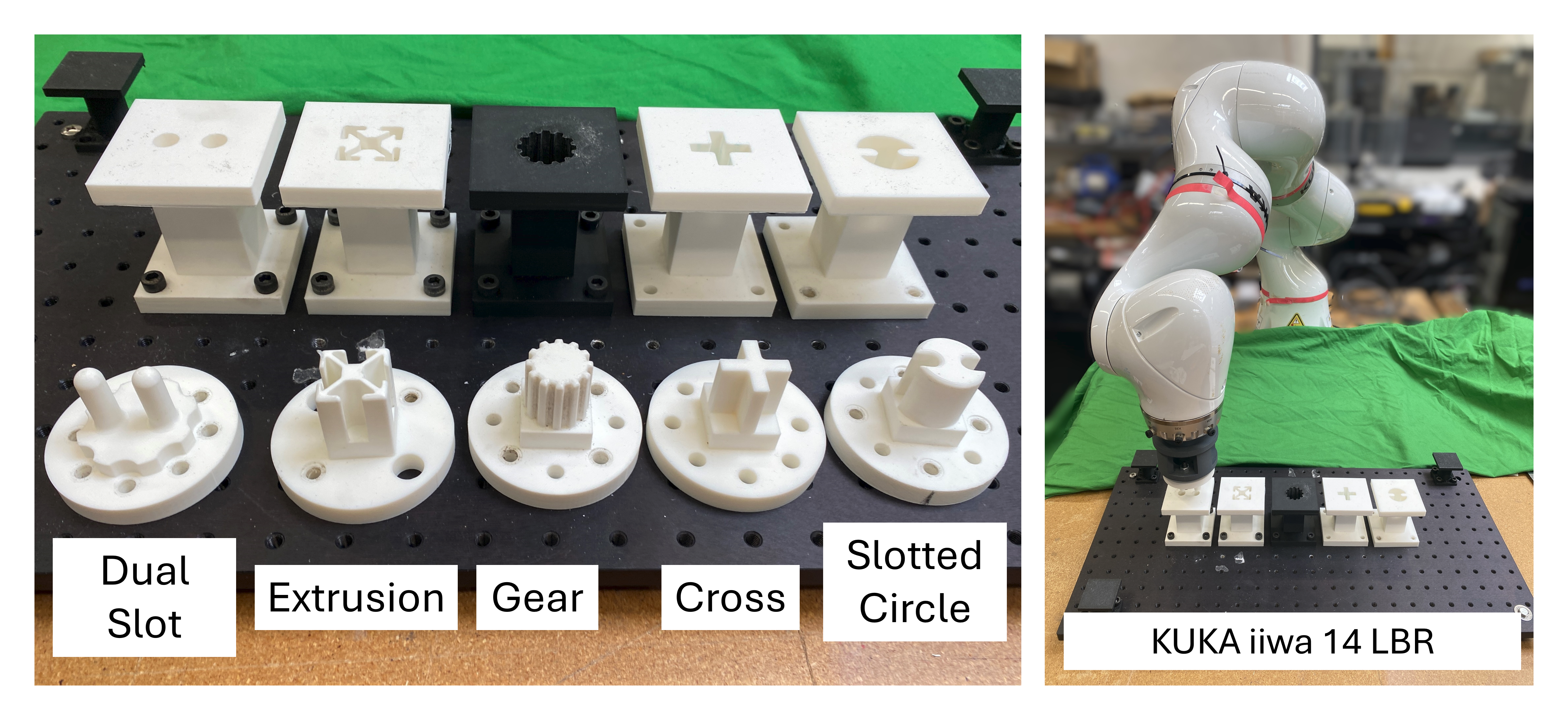}
    }
    \caption{Corresponding peg geometries (Left) and Experimental setup (Right). We select these non-convex geometries for their relevance to industrial assembly and 3D print each with maximum mating clearances ranging from 0.1 to 1.0 mm.}
    \label{fig:cell-setup}   
\end{figure}
\noindent Our experimental setup is illustrated in Fig. \ref{fig:cell-setup}. We examine five distinct peg-hole geometries selected based on their relevance in industrial assembly joints for complex parts and the non-convexity of their shapes. Each geometry adheres to established design and manufacturing guidelines for assembly \cite{Boothroyd2005}. To facilitate insertion, we incorporate guiding features such as filleted edges on the pegs. Additionally, we analyze the challenges posed by tight clearances, a primary factor contributing to jamming in assembly tasks. We set the clearance between the peg and hole between 0.1 and 1.0 mm. The entire setup is mounted on an optical breadboard to ensure stability and precise fixation. The peg is rigidly attached to the end-effector of a KUKA LBR iiwa 14 R820, a 7-DOF manipulator. The robot’s software stack provides an API with a built-in Cartesian impedance controller, allowing for the specification of stiffness and damping parameters. We tune these gains to achieve the desired compliance during assembly. Furthermore, the API enables dynamic switching between position control and impedance control modes, which we leverage at different stages of the operation. $\mathcal{M}$ is generated in a self-supervised manner in under 30 mins for a given geometry. For conducting our trials for pose estimation, we sample an initial pose offset from $U(-5mm,5mm)$ for $(x,y,z)$ and $U(-5^\circ, 5^\circ)$ for $(\alpha, \beta, \gamma)$ Success is defined by the threshold in Z of $\epsilon_{success} = 2mm$. 

\subsection{Results}\label{sec:results}

\begin{figure*}
    \centering
    \includegraphics[width=0.95\linewidth]{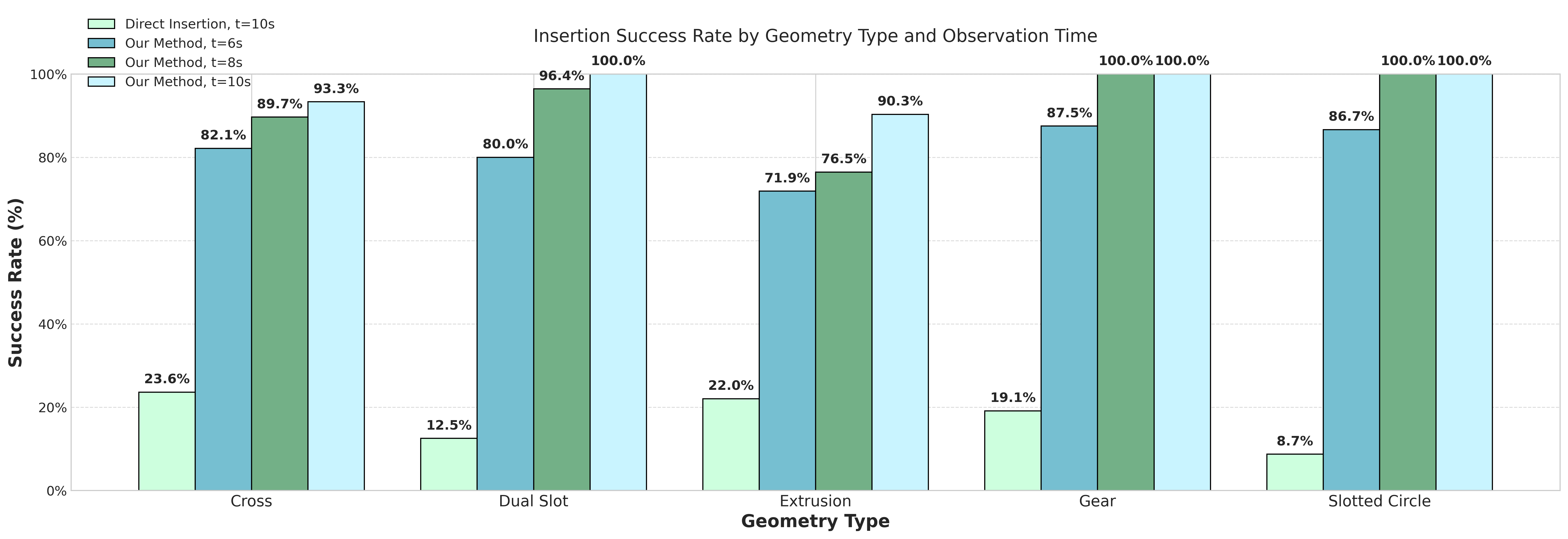}
    \caption{Success rates across geometry types, comparing our method ($t_{obs}$ = 6s, 8s, 10s) to direct insertion ($t_{obs}$ = 10s)}.
    \label{fig:success_results}
\end{figure*}

\begin{figure*}
    \centering
    \begin{minipage}{0.42\linewidth}
        \centering
        \includegraphics[width=\linewidth]{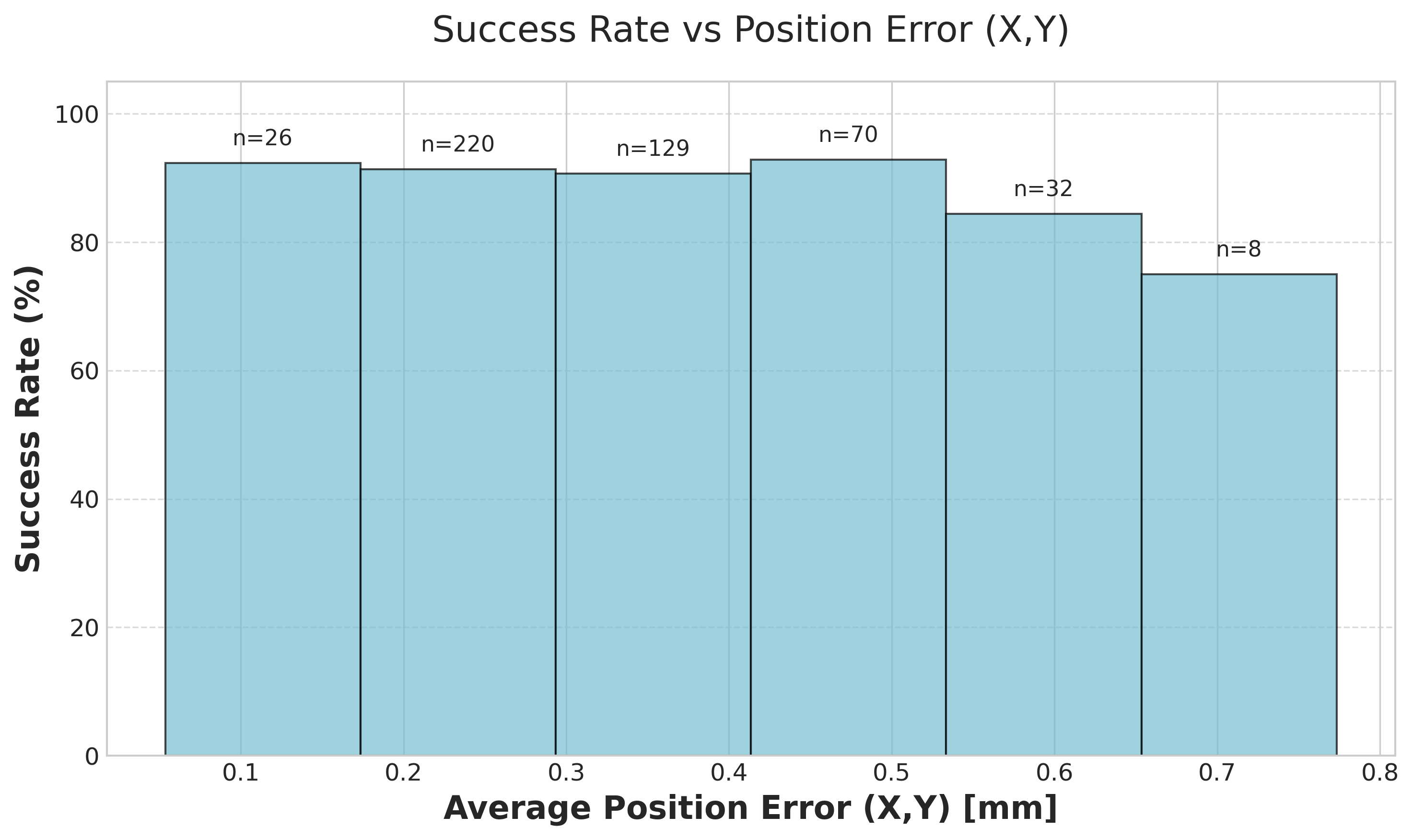}
    \end{minipage}
    \hfill
    \begin{minipage}{0.42\linewidth}
        \centering
        \includegraphics[width=\linewidth]{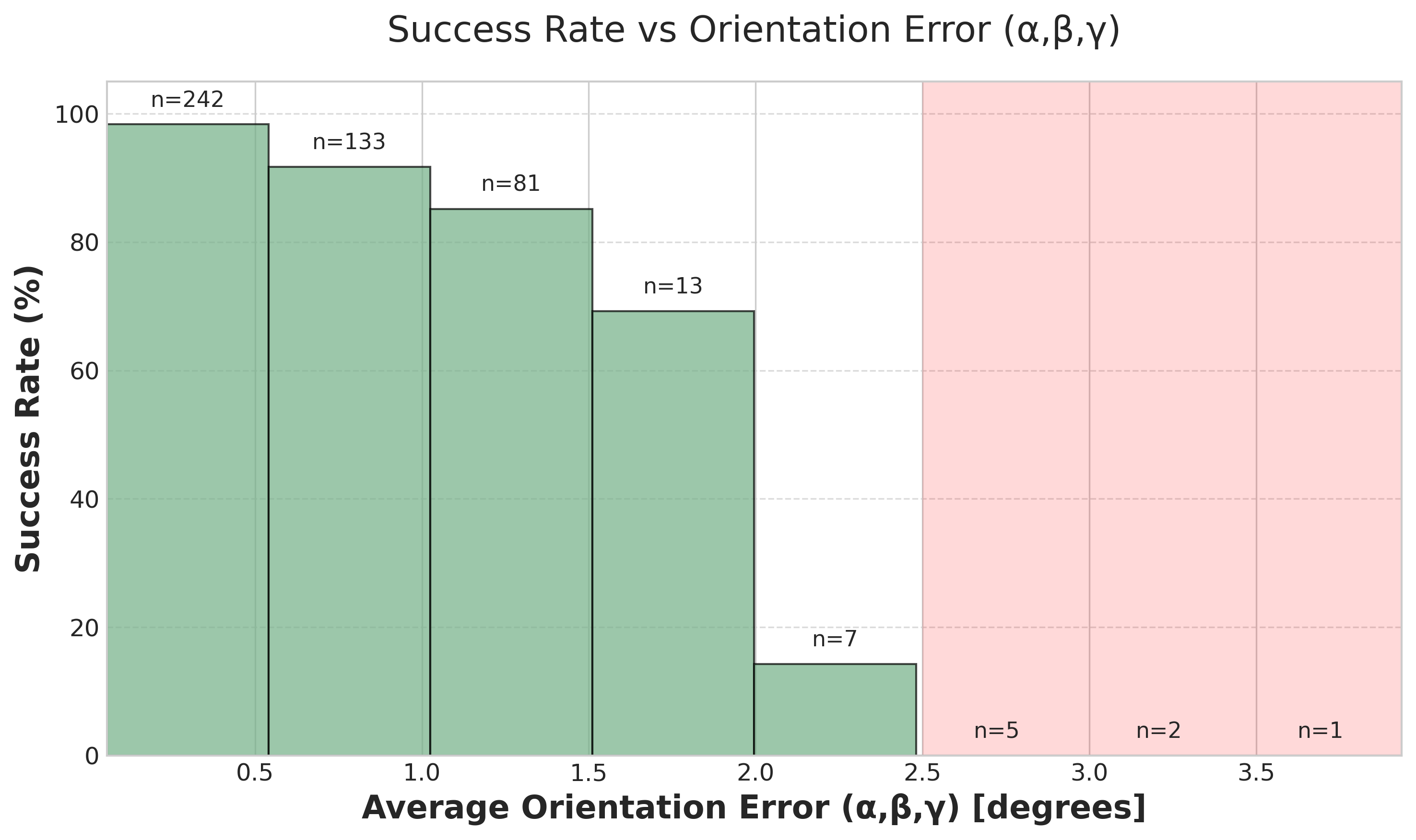}
    \end{minipage}
    \caption{Relationship between positioning accuracy and insertion success in robotic assembly tasks. The histograms illustrate how success rates vary with (left) positional error (X,Y axes, in millimeters) and (right) orientation error ($\alpha$, $\beta$, and $\gamma$, in degrees). Each bar represents the percentage of successful insertions within that error range, with sample sizes noted. The red-highlighted region in the orientation plot (>2.5°) indicates a critical threshold where success rates significantly decrease. Together, these results quantify the precision requirements for reliable assembly operations across multiple geometry types.}
    \label{fig:pose_results}
\end{figure*}

\begin{figure}
    \centering
    \includegraphics[width=0.96\linewidth]{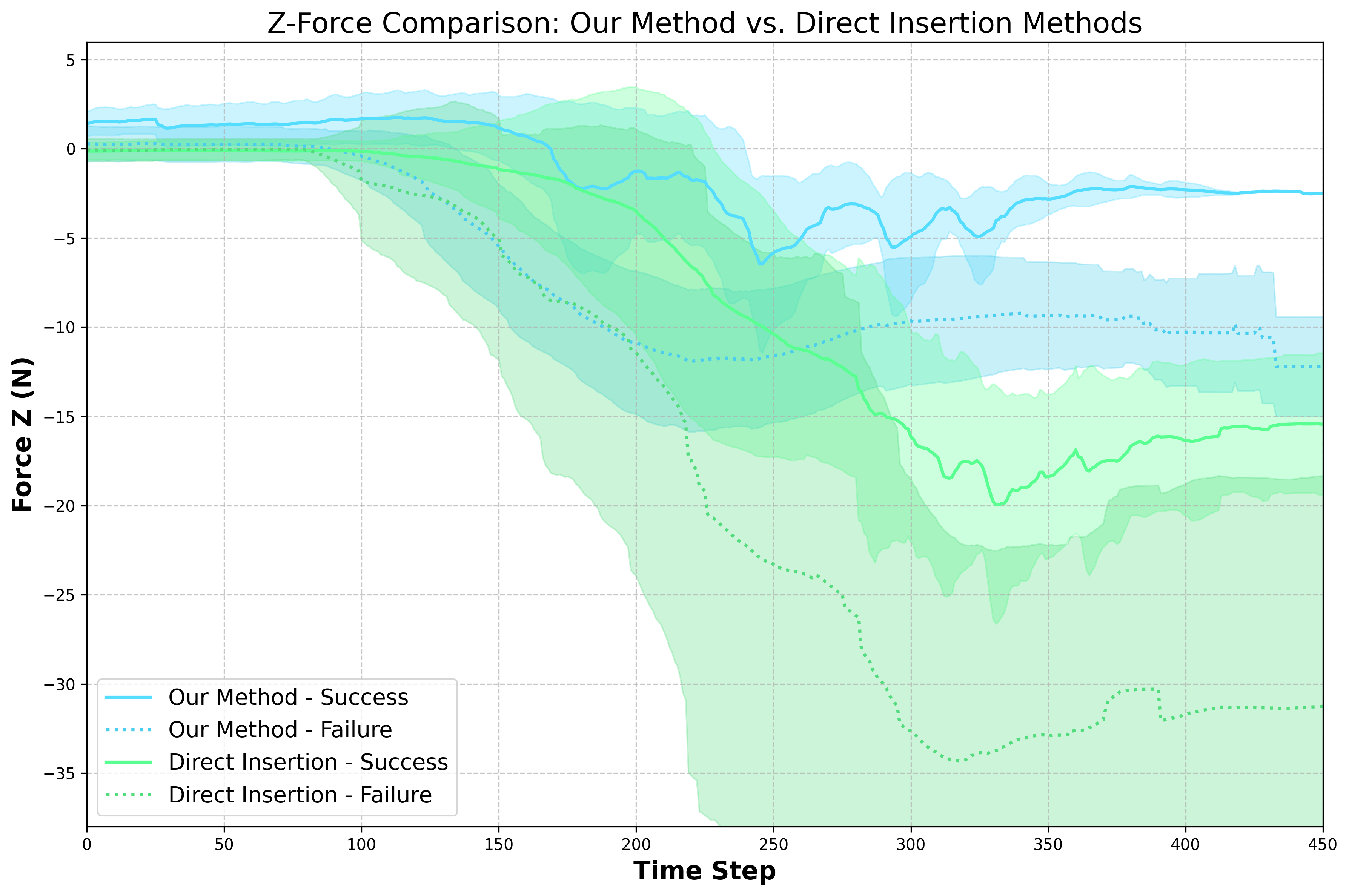}
    \caption{Z-axis force ($F_z$) profiles comparing standard and direct insertion methods in both successful and failed attempts. Shaded regions represent $\pm$ 1 standard deviation.}
    \label{fig:z_force}
\end{figure}

\noindent{\subsubsection{Pose Estimation} 
We perform a minimum of 25 trials for each of the 5 geometries (shown in Fig. \ref{fig:cell-setup}), for 3 different observation collection durations, totaling over 375 trials. Averaging across all geometries, we observed \textbf{mean absolute errors of 0.33, 0.32, and 0.33 mm in position (X, Y)} and \textbf{mean absolute errors of 0.85, 0.73, and 0.60 degree in orientation ($\boldsymbol{\alpha}$, $\boldsymbol{\beta}$, $\boldsymbol{\gamma}$)}, for observation times of 6, 8, and 10 seconds, respectively. Similarly, we observed \textbf{standard deviations in position error of 0.23, 0.20, and 0.19 mm} and \textbf{standard deviations in orientation error of 0.93, 0.78, and 0.47 degrees}, for observation times of 6, 8, and 10 seconds, respectively. Due to the exhaustive nearest-neighbor search and the iterative nature of the algorithm, the compute time is on the order of minutes - a faster algorithm is demonstrated in \cite{negi_icra}, in which a learned metric projection function reduces compute time by two orders of magnitude.} 

The orientation mean absolute error significantly decreases with longer observation collection time, while it remains constant for position mean absolute error. The standard deviation of the error decreases with longer observation time for both position and orientation. Even with a fixed number of observations after downsampling, longer observation time results in more accurate pose predictions due to the observations having higher variability. The higher variance leads to more accurate pose estimates by reducing the effects of biasing from measurement noise and local invariance. 

Registration of the contact manifold becomes more challenging along dimensions and in neighborhoods of lower variance - this issue is similarly found in ICP for point cloud registration \cite{censi_icp_2007}. This phenomenon is exemplified in the predicted values of Z, which has a mean absolute error of 2.62 mm across all trials. Moreover, the effects of invariance are also exhibited in longer observation time leading to more significant decrease in orientation error than position error - this is consistent with the contact manifold having higher variance along the orientation dimension than the position dimension. Nevertheless, the error in Z does not lead to misalignment and is easily compensated for by overshooting in the predicted Z direction; and the error in position is sufficiently low such that the success rate is only weakly correlated with the position error, as shown in Fig. \ref{fig:pose_results}. 

\noindent{\subsubsection{Insertion} We compare our method's insertion success rate against a direct insertion baseline, where the robot collects observations for 10 secs from partial insertion before attempting insertion along the peg's Z-axis under impedance control. This baseline is tested for 25 trials per geometry, with impedance parameters optimized for higher success.

As shown in Fig. \ref{fig:success_results}, the ``direct insertion'' method results in poor success rates, averaging 17.2\% across all geometries. Conversely, our method demonstrates high success rates of 81.6\%, 92.5\%, and 96.7\% across all geometries with observation times of 6, 8, and 10 seconds, respectively. The success rates of the ``direct insertion'' method highlight the insertion difficulty due to the geometry complexity and tight clearances, while the success rates of our method demonstrate how contact pose measurements can be used for precise pose estimation and significantly improve success rates on difficult assemblies. 

Insertion success rates improve with longer observation times due to increased pose estimation accuracy. The total time from partial to full insertion ranges from 7 to 11 seconds, primarily spent on observation. During this phase, the robot samples different poses while incrementally inserting the peg. The consistent trend of higher success rates with longer observation suggests that, given sufficient time, the method can achieve near 100\% success, as demonstrated on 3/5 geometries. Fig. \ref{fig:pose_results} shows that while success rate weakly correlates with position error, it strongly correlates with orientation error, indicating that failures are mostly due to higher estimation errors in orientation, which are likely due to lack of variability in the collected observations.

As shown in Fig. \ref{fig:z_force}, both the ``direct insertion method'' and our approach exhibit higher average insertion forces in failure cases due to jamming, where contact forces obstruct further insertion. Compared to direct insertion, our method reduces peak average insertion force by 68\% $(20.0 N \rightarrow 6.5 N)$ in success cases and 63\% $(34.0 N \rightarrow 12.5 N)$ in failure cases, significantly lowering the risk of damage to assembly parts. This highlights a key advantage of our method for safe and reliable robotic assembly.

\section{CONCLUSIONS}\label{sec:conclusions}

\noindent{We introduced a novel contact manifold-based method for precise pose estimation using only pose measurements and contact detection. Our approach constructs the manifold for complex, industrially relevant geometries and generates a contact sub-manifold online from contact pose observations. By registering this sub-manifold, we achieve sub-mm and sub-degree precision, increasing the success rate from 17.2\% to 96.7\% while reducing contact forces from 34.0 N to 12.5 N, effectively preventing jamming. Using only pose measurements and contact detection, our approach offers a robust and reliable solution for complex robotic assembly.} 

The presented methodology lays a foundation which can naturally be extended. Other parameters in the forward model, such as in-hand pose and kinematic parameters, can be estimated, as is demonstrated in \cite{negi_rss}. Observation collection can be made more efficient using active sensing techniques \cite{BehaviorSynthesisContactAware} \cite{hughes2024ergodic}. And an insertion controller can be designed based on the pose estimate and contact manifold.

\bibliographystyle{ieeetr}
\bibliography{references}

\end{document}